\pdfoutput=1
\documentclass[11pt]{article}

\usepackage[textsize=tiny, textwidth=0.5in, prependcaption]{todonotes}

\usepackage{microtype}
\usepackage{graphicx}
\usepackage{subfigure}
\usepackage{booktabs} % for professional tables

\usepackage{hyperref}

\usepackage{amsmath}
\usepackage{amssymb}
\usepackage{mathtools}
\usepackage{amsthm}

\usepackage[capitalize,noabbrev]{cleveref}

\crefformat{section}{#2\S#1#3}
\crefname{figure}{Fig.}{Figs.}
\crefname{table}{Tab.}{Tabs.}
\crefname{algorithm}{Algorithm}{Algorithms}
\usepackage{multirow}
\usepackage{tcolorbox}

\usepackage[preprint]{acl}

\usepackage{times}
\usepackage{latexsym}

\usepackage{multirow}  % For multirow cells in tables
\usepackage[T1]{fontenc}

\usepackage{colortbl}
\usepackage[utf8]{inputenc}

\usepackage{microtype}

\usepackage{inconsolata}
\usepackage{graphicx}
\usepackage{subcaption} % For subfigures
\usepackage{xspace}

\definecolor{californiagolden}{RGB}{255, 203, 5}

\usepackage{transparent}  % Add this for opacity
\usepackage[ruled,linesnumbered]{algorithm2e}
\usepackage{algpseudocode}
\usepackage{bm}
\usepackage{amsmath}
\usepackage{float}
\usepackage{caption}
\usepackage{graphicx} % Required for resizealgorithm

\reversemarginpar % Move notes to the left margin if needed
\newcommand{\allnotes}[1]{}
\renewcommand{\allnotes}[1]{#1} % Comment to turn off notes

\newcommand{\jx}[1]{\allnotes{\todo[color=blue!50]{jiarong: #1}}}

\newcommand{\frameworkname}{\textsc{$S^*$}\xspace}

\title{\frameworkname:~Test Time Scaling for Code Generation}

\author{
  Dacheng Li\thanks{Equal Contribution.}\thanks{Major Contributor.} \And
  Shiyi Cao$^{*\dagger}$ \And
  Chengkun Cao$^{\dagger}$ \And
  Xiuyu Li$^{\dagger}$ \AND
  Shangyin Tan \And
  Kurt Keutzer \And
  Jiarong Xing \And
  Joseph E. Gonzalez \And
  Ion Stoica \AND
  University of California, Berkeley
}

\begin{document}
\maketitle

\def\qwensevenb{Qwen2.5-Coder-7B-Instruct\xspace}
\def\qwenthirtytwob{Qwen2.5-Coder-32B-Instruct\xspace}
\def\fouromini{GPT-4o mini\xspace}

\newcommand{\myparagraph}[1]{\vspace{0pt}\paragraph{#1}}
\newcommand{\sect}[1]{Section~\ref{#1}}
\newcommand{\ssect}[1]{\S~\ref{#1}}
\definecolor{lightblue}{rgb}{0.90, 0.95, 0.99}

\begin{abstract}
%This document provides a basic paper template and submission guidelines.
% Abstracts must be a single paragraph, ideally between 4--6 sentences long.
% Gross violations will trigger corrections at the camera-ready phase.
Increasing test-time compute for LLMs shows promise across domains but remains underexplored in code generation, despite extensive study in math. In this paper, we propose \frameworkname, the first hybrid test-time scaling framework that substantially improves the coverage and selection accuracy of generated code. \frameworkname extends the existing parallel scaling paradigm with sequential scaling to push performance boundaries. It further leverages a novel selection mechanism that adaptively generates distinguishing inputs for pairwise comparison, combined with execution-grounded information to robustly identify correct solutions.
% combining LLM-based evaluation with execution-grounded verification to robustly identify correct solutions.
% Increasing the test-time compute of large language models (LLMs) has shown promising results. While extensive studies have been conducted in the math domain, the counterpart in code generation has been less explored. In this paper, we propose a simple and effective two-stage approach we call ~\frameworkname that improves (1) current repeated sampling paradigm by integrating revision from code execution feedback to improve the quality of each code samples, and (2) current majority-voting selection based method by grounding LLM-based selection method with code execution results.
% LLM-as-a-judge paradigm with test case execution grounding. 

%\kurt{as I'll bring up later, I think the consistent/uniform improvements across models is more impressive than the smaller models surpass larger ones 
%- unless you point to a new application that the smaller models can now handle that they could not before.}
We evaluate across 12 Large Language Models and Large Reasoning Model and show: (1) \frameworkname consistently improves performance across model families and sizes, enabling a 3B model to outperform GPT-4o-mini; (2) \frameworkname  enables non-reasoning models to surpass reasoning models—GPT-4o-mini with \frameworkname outperforms o1-preview by 3.7\% on LiveCodeBench; (3) \frameworkname 
further boosts state-of-the-art reasoning models—DeepSeek-R1-Distill-Qwen-32B with \frameworkname achieves 85.7\% on LiveCodeBench, approaching o1 (high) at 88.5\%.  
Code will be available under \url{https://github.com/NovaSky-AI/SkyThought}.
\end{abstract}
\section{Introduction}
Increasing test-time compute has emerged as a powerful approach for improving the performance of large language models (LLMs) across diverse tasks~\citep{openai_learning_to_reason_2024, guo2025deepseek,qwen_qwq_2024,muennighoff2025s1,sky_t1_2025,brown2024large,snell2024scaling}. In particular, test-time scaling has been extensively explored in mathematical reasoning, where parallel sampling increases solution coverage, sequential refinement improves individual samples through rethinking and revising, and reward models guide the search process more effectively~\citep{ehrlich2025codemonkeys, snell2024scaling, li2024rethinkmcts}. These methods collectively push the performance boundaries of LLMs by leveraging additional compute during inference.
%\dacheng{@Matei: cost comparison: small + test > large pass@1}
\begin{figure}[!t]
    \centering
    % First subfigure
    \includegraphics[width=0.48\textwidth]{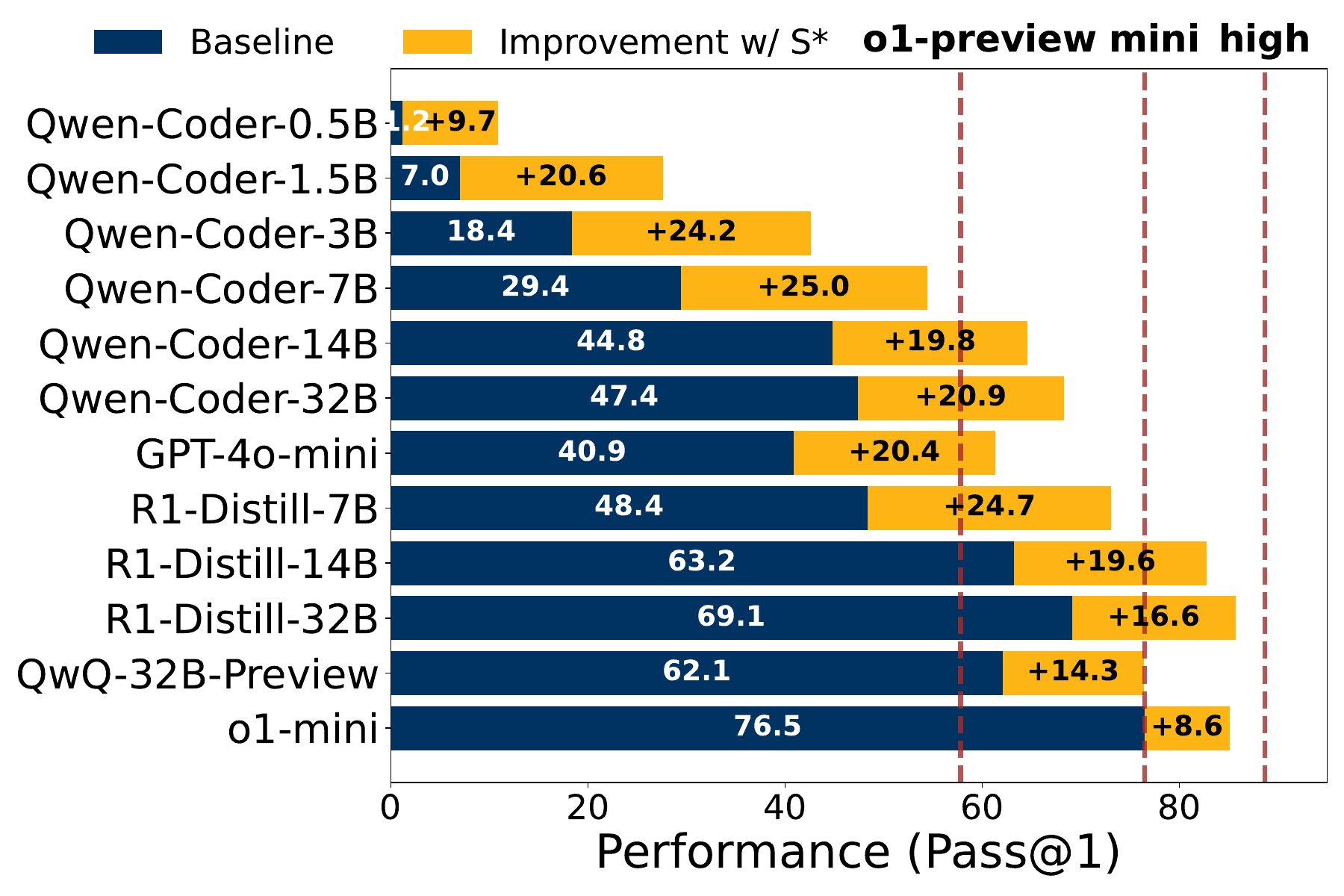}      \caption{\textbf{Performance improvement with~\frameworkname in LiveCodeBench (v2)}~\citep{jain2024livecodebench}. \frameworkname consistently improves models across different sizes, allowing non-reasoning models to surpass reasoning models and open models to be competitive with o1 (high reasoning effort). "Qwen-Coder" denotes "Qwen2.5-Coder-Instruct,"~\citep{hui2024qwen2} and "R1-Distill" denotes "DeepSeek-R1-Distill-Qwen." ~\citep{guo2025deepseek}. }%.\dacheng{run o1-high}}
    \label{fig:all_models_performance}
    \vspace{-5mm}
\end{figure}

\begin{figure*}[!t]
    \centering
    \includegraphics[width=1\linewidth]{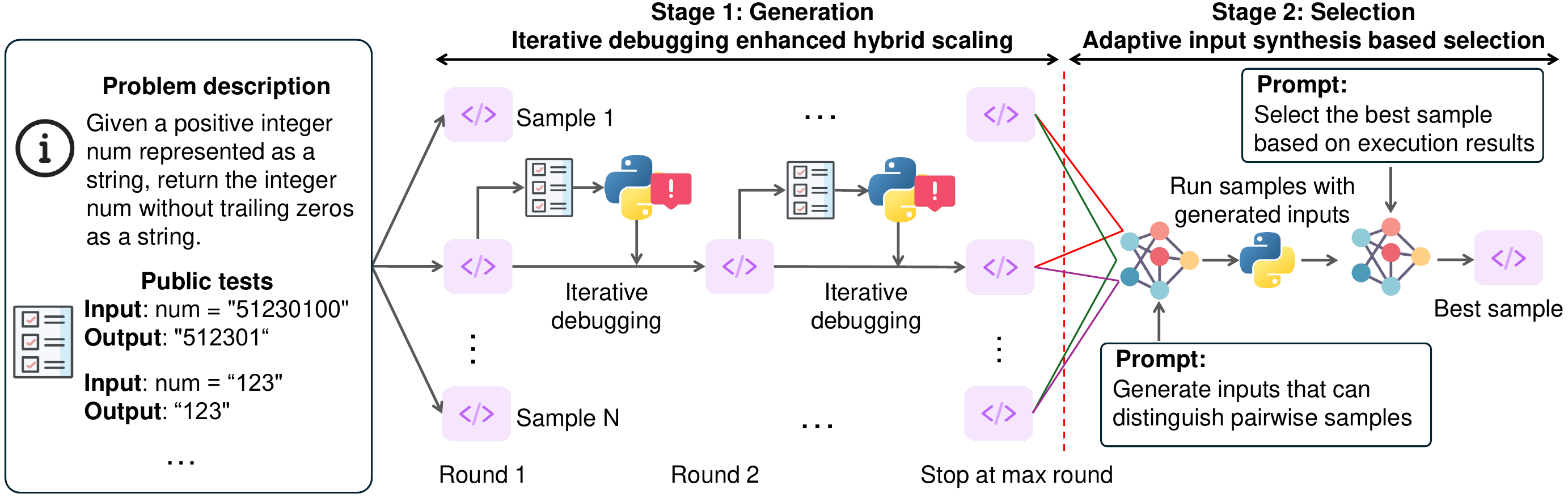}
\caption{
\textbf{Overview of \frameworkname}. \textbf{Stage 1: Generation}---\frameworkname enhances parallel samples through iterative debugging. Each sample is tested using public test cases executed via an interpreter, with outputs and/or error messages used to guide the next round of sample generation.
\textbf{Stage 2: Selection}---\frameworkname selects the best sample by prompting an LLM to generate inputs that differentiate between paired samples, then leveraging actual execution results to inform the LLM to determine the optimal choice.
% \frameworkname builds on the paradigm of a combination of revision and parallel samples. It first performs a self-refinement on the initial solution to improve algorithm efficiency, and subsequent self-debug round to correct solutions based on public test execution feedback. At the end, it undergoes a selection policy module to select the best solution to output.
}
\vspace{-5mm}
\label{fig:Design}
\end{figure*}

Despite these advancements in the math domain, the potential of test-time scaling for code generation---a domain with both fundamental importance and widespread practical applications---remains under-explored. Code generation introduces unique challenges compared to math reasoning. Correctness in math can often be verified through rule-based string matching with reference answers~\citep{guo2025deepseek, AceCoder}, whereas validating code requires executing a large set of test cases to accurately check functional correctness~\citep{liu2023your}. This dependence on execution increases the complexity of test-time scaling and complicates the design of reward models~\citep{AceCoder}. However, code generation also offers a distinct advantage: The availability of programmatic interpreters enables the execution of programs to obtain precise outputs and error messages, providing a reliable grounding mechanism for improving generation and selection~\citep{chen2023teaching, li2022competition}.

In this paper, we propose \frameworkname, the first hybrid test-time scaling framework for code generation, which substantially improves both coverage \footnote{The fraction of problems that are solved by any generated
sample~\citep{brown2024large}.} and selection accuracy. \frameworkname pushes the limits of existing parallel scaling strategies by integrating sequential scaling through \emph{iterative debugging}, while introducing a novel adaptive selection mechanism grounded in execution. The framework operates in two key stages, as shown in \cref{fig:Design}.

\textbf{First}, in the generation stage, \frameworkname augments parallel sampling~\citep{brown2024large, li2022competition} with sequential scaling via iterative debugging. Each generated sample is executed on public test cases to obtain outputs and/or error messages, which are fed back into the model to iteratively refine the code. 
% Notably, we find that even models capable of hypothetical debugging during reasoning benefit significantly from this execution-grounded feedback loop, leading to substantial performance gains.
\textbf{Second}, in the selection stage, existing methods often rely on generating test inputs indiscriminately, which can fail to effectively differentiate between candidate solutions~\citep{chen2022codet, AceCoder}. To overcome this limitation, \frameworkname introduces \textit{adaptive input synthesis}: for each pair of samples, an LLM is prompted to generate distinguishing test inputs. These inputs are executed, where the outputs are further provided to ground the LLM to select the best sample.
%and the sample that performs best on the synthesized inputs is selected as the final output. 
This adaptive, execution-grounded approach ensures robust identification of correct solutions (\cref{sec:ablation_selection}).

\begin{figure}[!t]
    \centering
    % First subfigure
    \includegraphics[width=1.0\linewidth]{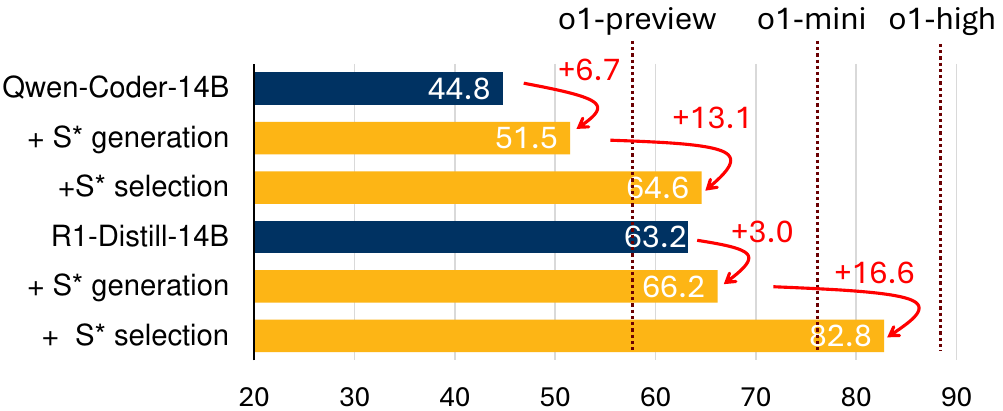}      
    \caption{
    % \textbf{Example overview of performance.}
    \textbf{Ablation of \frameworkname performance benefits}: Qwen2.5-Coder-14B-Instruct (denoted as Qwen-Coder-14B)~\citep{hui2024qwen2} with \frameworkname can surpass o1-preview without \frameworkname. DeepSeek-R1-Distill-Qwen-14B (denoted as R1-Distill-14B)~\citep{guo2025deepseek} with \frameworkname outperforms o1-mini without \frameworkname.
    % \xiuyu{should we use +xx\% instead than +xx $\times$}}
    }
    % \jx{We can potentially replace the non-reasoning model to beat o1-preview.}
    \label{fig:breakdown}
\end{figure}

\frameworkname is a general approach that outperforms zero-shot generation and existing test-time scaling methods. We evaluate \frameworkname on 12 models, spanning a wide range of sizes, both open and closed, instruction-based and reasoning models. \frameworkname consistently enhances performance across these diverse settings. Notably, \frameworkname enables: (1) Small models to surpass larger models within the same family: Qwen2.5-7B-Instruct + \frameworkname outperforms Qwen2.5-32B-Instruct on LiveCodeBench by 10.7\%; (2) Instruction-based models to outperform reasoning models: GPT-4o-mini + \frameworkname surpasses o1-preview by 3.7\%; and (3) Open reasoning models to achieve performance competitive with state-of-the-art closed models: DeepSeek-R1-Distill-Qwen-32B + \frameworkname achieves 85.7\% on LiveCodeBench, approaching the state-of-the-art performance of o1-high at 88.7\%.
\cref{fig:breakdown} provides an overview of the performance improvements enabled by our techniques. %We further conduct comprehensive ablation studies on variants of iterative debugging and selection methods, as detailed in~\cref{sec:exp}. 
In summary, our contributions are:

% We evaluate\shiyi{I am here} \frameworkname on 12 models across model sizes, open and closed, and both instruction-based or reasoning models. \frameworkname has shown consistent improvement over zero-shot generation or existing popular test-time scaling methods. In particular, \frameworkname~allows (1) small models to outperform large models in the same family: Qwen2.5-7B-Instruct + \frameworkname outperforms Qwen2.5-32B-Instruct on LiveCodeBench by 10.7\%; (2) instruction-based model to outperform reasoning models: GPT-4o-mini + \frameworkname outperforms o1-preview by 3.7\%; 
% and (3) open reasoning models to perform competitively to the state-of-the-art: DeepSeek-R1-Distill-Qwen-32B + \frameworkname~ achieves 85.7\% on LiveCodeBench, where the state-of-the-art (o1-high) is 88.7\%.
% Figure~\ref{fig:breakdown} shows an example overview of the benefits of our proposed techniques.
% We further include comprehensive ablation studies on variants of iterative debugging and selection methods in the experiment section~\ref{sec:exp}.
\begin{enumerate}
\item We propose \frameworkname, the first hybrid test-time scaling framework for code generation, which augments parallel scaling with sequential scaling via iterative debugging and introduces adaptive test input synthesis using LLMs for robust sample selection.
\item We conduct extensive evaluations on LiveCodeBench and CodeContests, demonstrating that \frameworkname consistently improves performance across diverse model families and sizes.
\item We will release all software artifacts, model generations, and intermediate results to support and accelerate future research in this area.
\end{enumerate}

\section{Related work}
\myparagraph{Test Time Scaling for LLMs.} 
Existing approaches to increase test-time compute can be broadly categorized into two paradigms: parallel scaling and sequential scaling~\citep{muennighoff2025s1}. 
% Existing approaches for increasing test-time compute can be generally classified into two basic directions: parallel and sequential. 
Parallel scaling (i.e., repeated sampling) involves generating multiple solutions simultaneously and selecting the best one, a strategy commonly known as Best-of-N.
Coverage---the fraction of problems solved by any of these N samples---continues to improve as $N$ increases~\citep{chollet2019measure, irvine2023rewarding}, even at the scale of $10^4$ to $10^6$~\citep{brown2024large, li2022competition}. 
However, common selection strategies, such as (weighted) majority voting~\citep{wang2022self} and reward model scoring~\citep{christiano2017deep, lightman2023let, wang2024math, wu2024inference, beeching2024dvts, pan2024swegym}, often struggle to select the correct best sample in parallel scaling~\citep{brown2024large, hassid2024larger, stroebl2024inference}. In this paper, we propose a novel method that improves selection for coding tasks. 
% Although increases the coverage, it is often difficult to select the correct best sample in parallel scaling~\citep{brown2024large, hassid2024larger, stroebl2024inference}.

Sequential scaling, on the other hand, encourages the model to refine its reasoning over multiple steps. This includes methods like chain-of-thought (CoT) prompting~\citep{wei2022chain, nye2021show}, and iterative rethinking and revision~\citep{madaan2024self, lee2025evolving, hou2025advancing, huang2022large, min2024imitate, sky_t1_2025, muennighoff2025s1, wang2024theoretical, li2025llms}. Noticeably, OpenAI o1, DeepSeek R1, Qwen QwQ, and Kimi employ in-context long CoT with revision and backtracking to find the best solution~\citep{openai_learning_to_reason_2024, guo2025deepseek, qwen_qwq_2024, team2025kimi}. In this paper, we leverage iterative debugging from test execution feedback for sequential scaling code generation performance~\citep{chen2023teaching}.

% On the other hand, sequential scaling encourages the model to produce intermediate steps, e.g., chain-of-thoughts (CoT) and scratchpad~\citep{wei2022chain, nye2021show}, or to rethink and revise a previous solution~\citep{madaan2024self,lee2025evolving,hou2025advancing, huang2022large, min2024imitate, sky_t1_2025, muennighoff2025s1, wang2024theoretical}. Noticeably, OpenAI o1, DeepSeek R1, Qwen QwQ, and Kimi employ in-context long CoT with revision and backtracking to find the best solution~\citep{openai_learning_to_reason_2024, deepseek_r1_lite_2024, guo2025deepseek, qwen_qwq_2024, team2025kimi}. 

% Our work is unique in that it adopts a hybrid approach, integrating both parallel and sequential scaling while enhancing them with iterative debugging and adaptive input selection.
% Our work adopts a hybrid approach, integrating both parallel and sequential scaling.

\myparagraph{Test Time Scaling for Code Generation.}
~\citet{chen2022codet, huang2023enhancing, jiao2024preference} use models to generate code samples and test cases, selecting the final sample in a self-consistency manner~\citep{wang2022self, AceCoder}. However, these approaches often suffer from model hallucination, where the model fails to accurately predict the output of a test input~\citep{jain2024livecodebench, AceCoder,gu2024cruxeval}.
%CodeT uses RANSAC algorithm to select the correct sample using model-generated test cases to filter model-generated solutions~\citep{chen2022codet}. 
AlphaCode explores large-scale parallel sampling with a trained model to generate test cases for filtering and selection~\citep{li2022competition}. AlphaCodium uses a series of self-revision on both public demonstration and model-generated tests to improve solutions~\citep{ridnik2024code}. ~\citet{saad2024archon} searches over various inference techniques and finds that parallel sampling with model-generated tests works well for CodeContests problems~\citep{li2022competition}. Unlike methods relying solely on parallel sampling or sequential scaling, we use a hybrid approach that combines their advantages.
% Our framework is different than existing approach in that it leverages parallel scaling, sequential scaling, and a hybrid method to leverage both model-generated test inputs and LLM-as-a-judge approach to do selection.

% \frameworkname adopts a hybrid test-time scaling approach for code generation by extending parallel scaling with sequential scaling through iterative debugging for increased coverage and employing adaptive input synthesis to improve selection accuracy.

% Our framework differs by enhancing sequential scaling with execution-grounded iterative debugging and leveraging adaptive input synthesis to distinguish samples and select the best one.

\myparagraph{Hybrid Test-Time Scaling.}
Many works in the math domain study hybrid approaches that combine parallel and sequential scaling, often leveraging reward-model-guided tree search algorithms, such as Monte-Carlo Tree Search (MCTS), to effectively navigate the solution space~\citep{gao2024interpretable, li2024rethinkmcts, silver2016mastering, snell2024scaling, hendrycks2021measuring}. S1~\citep{muennighoff2025s1} primarily focuses on sequential scaling but observes diminishing returns and thus incorporates parallel-based approaches like majority voting and tree search to further enhance performance.

In contrast, our work applies hybrid scaling to code generation tasks without relying on tree search methods, as developing a general and effective reward model for the code generation domain remains challenging~\citep{AceCoder}. Instead, \frameworkname augments parallel scaling with sequential scaling via execution-grounded iterative debugging to improve coverage and introduces adaptive input synthesis to enhance selection accuracy.

% \myparagraph{Hybrid-based Test Time Scaling.}
% S1~\citep{muennighoff2025s1} primarily studies sequential scaling but finds that its effectiveness eventually plateaus. To address this, it explores parallel-based approaches, including majority voting and tree searching. 
% There are also hybrid approaches leveraging tree search-like algorithms, such as using Monte-Carlo Tree search~\citep{gao2024interpretable,li2024rethinkmcts, silver2016mastering}. 
% ~\citet{snell2024scaling} studies the combination of parallel scaling and sequential scaling on a math benchmark~\citep{hendrycks2021measuring}. 
% Our work combines parallel scaling and sequential scaling for code generation without relying on tree search methods, as developing a general and effective reward model for code generation domain remains challenging~\citep{AceCoder}.

% reward models for coding remain underdeveloped~\citep{AceCoder}.

\myparagraph{Concurrent Work.} 
CodeMonkeys is a noticeable concurrent work to this paper, released on Arxiv in Jan 2025~\citep{ehrlich2025codemonkeys}. It also generates multiple samples in parallel and revises samples. However, CodeMonkeys focuses on the software engineering domain, optimizing performance on SWE-Bench (Chowdhury et al., 2024),
which addresses challenges such as identifying files
that need to be edited. In contrast, our work focuses on competition-level code generation, where domain differences influence our algorithm choice. For instance, during sequential scaling, CodeMonkeys requires a model to generate tests over multiple iterations, while we instead incorporate feedback from public tests (ablated variants in~\cref{ssect:ablation_selfdebug}).

\section{Method}
\label{sec:method}

% In this section, we introduce the \frameworkname framework. 
\frameworkname takes as input a coding problem $\mathbf{P}$ and a code generation model $\mathbf{M}$. The model $\mathbf{M}$ aims to generate a program solution $\mathbf{X(\cdot)}$ that maps inputs to outputs according to the problem specification.

We adopt the standard setup widely used in existing coding benchmarks~\citep{chen2021evaluating, li2022competition, li2023taco, jain2024livecodebench, hendrycksapps2021, gulwani4foundations}. Each coding problem $\mathbf{P}$ consists of a natural language description and a set of public and private test cases, each represented as input-output pairs. 

Private tests evaluate the correctness of $\mathbf{X}$ but remain inaccessible to $\mathbf{M}$ during code generation. A solution is considered correct if it passes all private tests. In contrast, public tests are provided to clarify the problem's intent and are typically included in the prompt. Public tests are usually far fewer than private tests; for instance, in CodeContests~\citep{li2022competition}, there are, on average, 2.0 public tests and 202.1 private tests per problem. This contrasts with mathematical reasoning tasks, where evaluation typically relies on exact string matching of the final solution without additional test information~\citep{li2024numinamath}.

\subsection{The \frameworkname Framework}
\frameworkname is a two-stage hybrid test-time scaling framework consisting of \emph{Generation} and \emph{Selection} stages, as demonstrated in~\cref{fig:Design}. It extends parallel sampling with sequential sampling via iterative debugging to \emph{improve coverage} and employs adaptive input synthesis during selection to \emph{enhance selection accuracy}, leveraging execution results throughout the process. An example of effect for different stages can be found in~\cref{fig:breakdown}.

\paragraph{Stage 1: Generation.}
\label{sec:scale_parallel_samples}
In the generation stage, \frameworkname improves coverage by extending parallel scaling with sequential scaling through \textit{iterative debugging grounded with execution feedback}. Specifically, \frameworkname first generates $\mathbf{N}$ initial samples independently, leveraging parallel sampling techniques~\citep{chen2023teaching}. Each sample is then refined through up to $\mathbf{R}$ rounds of sequential revision, informed by execution results on public test cases. The revision process halts once a sample passes all public tests or reaches the maximum number of revision attempts.
% \shiyi{We also experimented with self-generated tests for additional debugging rounds but observed no further performance gains (see~\cref{ssect:ablation_selfdebug}) is removed from here for cleaner message.}
% We also experimented with self-generated tests for additional debugging rounds but observed no further performance gains (see~\cref{ssect:ablation_selfdebug}).

\paragraph{Stage 2: Selection.}
\label{sec:scale_selection}
After generating $\mathbf{N}$ candidate solutions, the next step is to identify the best one. Since the public tests are already used during the generation stage, additional evaluation is needed for reliable selection. We investigate two existing approaches: (1) LLM-as-a-judge~\citep{zheng2023judging}, which relies on pre-trained knowledge to compare candidate solutions, and (2) generated test cases~\citep{li2022competition, chen2022codet}
%\shiyi{add citation to codeT}, 
which uses synthesized test cases to guide selection.

Unfortunately, we find that LLM-based judging alone often struggles to predict program behavior accurately, while generated tests frequently fail to provide reliable outputs for grounding the selection or to produce high-quality inputs that effectively distinguish samples (see \cref{tab:diff_selection}).

% Unfortunately, we find that LLM-based judging alone often struggles to predict program behavior accurately, while generated tests frequently fail to produce high-quality inputs that effectively distinguish samples (see \cref{tab:diff_selection}).

\begin{algorithm}[t]
    \footnotesize
	\caption{Best Sample Selection in \frameworkname}
    \label{alg:selection1}

	\KwIn{Problem description: $P$}
	\KwIn{Candidate samples: $X$}
	\KwOut{The best selected sample: $x^*$}
	
	% Generate initial test inputs using LLM
	$\mathcal{T} \gets \texttt{llm\_test\_input\_gen($P$)}$ \label{line:execute}

    % \tcc{Execute samples with python interpreter}
    
	% Execute all samples on test inputs
	$\mathcal{O} \gets \texttt{sample\_execution}(X, \mathcal{T})$
	
	% Cluster samples based on execution results
	$\mathcal{C} \gets \texttt{sample\_clustering}(\mathcal{O})$ \label{line:cluster}
	
	% Initialize cluster scores
	$\text{Scores} \gets \mathbf{0}$ 
    
	% \tcc{Using adaptive input to distinguish clusters}
    
	% Adaptive evaluation across clusters
	\For{\textbf{each} pair $(C_i, C_j) \in \mathcal{C}$}{

        Sample $x_i$, $x_j$ from $C_i$, $C_j$
        
        $\mathcal{T}_{\text{a}} \gets \texttt{adaptive\_input\_gen}(x_i, x_j)$ \label{line:adaptive}

	    better\_idx = $\texttt{exec\_and\_llm\_select}(x_i, x_j, \mathcal{T}_{\text{a}})$
	    
	    $\text{Scores}$[better\_idx] += 1 \label{line:increase}
	}
	
	% Select the best-performing cluster
	$C^* \gets \arg\max (\text{Scores})$
	
	% Randomly pick a final sample from the best cluster
	$x^* \gets \texttt{random\_pick}(C^*)$
	
	\Return $x^*$
\end{algorithm}

To overcome these limitations, \frameworkname introduces \emph{adaptive input synthesis}, a hybrid selection approach that integrates LLM evaluation with execution-grounded verification, as illustrated in Algorithm~\ref{alg:selection1}. First, we prompt an LLM to synthesize a set of test inputs. We execute these inputs and cluster the $\mathbf{N}$ samples based on their execution outputs (Line~\ref{line:execute} to Line~\ref{line:cluster})~\citep{li2022competition}. We then perform pairwise comparisons across clusters: for each comparison, we prompt the LLM to generate distinguishing inputs, execute both samples using these inputs, and select the superior one based on the execution results (Line~\ref{line:adaptive} to Line~\ref{line:increase}). This adaptive selection process grounds LLM evaluations in concrete execution feedback, resulting in more reliable and accurate sample selection compared to naive LLM judging or generated tests-based methods (see~\cref{sec:exp}).

\begin{table*}[!t]
\centering
\renewcommand{\arraystretch}{1.2}
\setlength{\tabcolsep}{6pt}
\resizebox{0.9\textwidth}{!}{%
\begin{tabular}{l cccccc c ccc c c}
\toprule
\textbf{Method} 
& \multicolumn{6}{c}{\textbf{Qwen2.5-Coder-Instruct}} 
& \textbf{4o-mini} 
& \multicolumn{3}{c}{\textbf{R1-Distill}} 
& \textbf{QwQ} 
& \textbf{o1-mini} \\
\cline{2-7} \cline{9-11}
& \textbf{0.5B} & \textbf{1.5B} & \textbf{3B} & \textbf{7B} & \textbf{14B} & \textbf{32B}
& 
& \textbf{7B} & \textbf{14B} & \textbf{32B}
& 
& \\
\midrule
\textbf{Zero-Shot}      
& 1.2  & 7.0  & 18.4 & 29.4 & 44.8 & 47.4 
& 40.9 
& 48.4 & 63.2 & 69.1 
& 62.1 
& 76.5 \\
\textbf{Majority Vote}  
& 2.5  & 11.0 & 25.2 & 40.5 & 50.8 & 55.9 
& 46.6 
& 58.7 & 68.1 & 75.8 
& 67.3 
& 81.6 \\
\textbf{Self-Debugging} 
& 2.4  & 9.4  & 27.8 & 39.9 & 51.5 & 59.5 
& 51.7 
& 58.4 & 66.2 & 70.1 
& 59.3 
& 79.9 \\
\textbf{S*}             
& \textbf{10.9} & \textbf{27.6} & \textbf{42.7} & \textbf{54.4} & \textbf{64.6} & \textbf{70.1} 
& \textbf{61.3} 
& \textbf{73.2} & \textbf{82.8} & \textbf{85.7} 
& \textbf{76.3}
& \textbf{85.3} \\
\bottomrule
\end{tabular}%
}
\caption{\textbf{Pass@1 of zero-shot, majority voting~\citep{wang2022self, li2022competition}, self-debugging~\citep{chen2023teaching}, and \frameworkname on LiveCodeBench (v2)}. Bold text denotes the best performance. "R1-Distill", "QwQ", "4o-mini" is short for "DeepSeek-R1-Distill-Qwen"~\citep{guo2025deepseek}, "QwQ-32B-Preview"~\citep{qwen_qwq_2024}, and "GPT-4o-mini"~\citep{achiam2023gpt} respectively. \textit{\frameworkname consistently outperforms other baselines.}}\label{tab:exp_diff_strategies}
%\vspace{-5mm}
\end{table*}
\section{Evaluation}
\label{sec:exp}
%\joey{Rather than outline the section, you could start with the goals.  Something like:
%We evaluate our approach on a wide range of models and model sizes and show:
%\begin{enumerate}
%    \item \frameworkname consistently results in a NUMBER improvement. (subsection ref)
%    \item Some ablation finding 1 (subsection ref)
%    \item some ablation finding 2 (subsection ref)
%\end{enumerate}
%This typically helps setup the reader for what you are %claiming (expect to show).
%}
In this section, we evaluate \frameworkname across a diverse set of instruction-based and reasoning models, spanning various model families, sizes, and access types (open and closed), as well as multiple benchmarks~\citep{jain2024livecodebench, li2022competition}.

Our key findings demonstrate the generality and effectiveness of \frameworkname:
\begin{enumerate}
\item \frameworkname consistently improves model performance across different families, sizes, and types, and generalizes effectively to multiple code generation benchmarks, including LiveCodeBench (\cref{sec:exp_main}) and CodeContests (\cref{sec:exp_other_benchmark}), showcasing its robustness and broad applicability.
\item \frameworkname outperforms existing widely-used test-time scaling methods, including self-debugging~\citep{chen2023teaching} and majority voting~\citep{wang2022self, li2022competition}, by enhancing both coverage and selection accuracy (\cref{sec:exp_compare_methods}).
% \item \frameworkname improves performance not only on LiveCodeBench but also generalizes effectively to other code generation benchmarks, highlighting its robustness (\cref{sec:exp_other_benchmark}).
\end{enumerate}
% In this section, we evaluate \frameworkname across a wide range of instruction-based and reasoning-based models, as well as benchmarks~\citep{jain2024livecodebench, li2022competition}. 
% We have the following key findings:
% \begin{enumerate}
%     \item \frameworkname consistently improves model capabilities, across different model families and sizes (\cref{sec:exp_main}).
%     \item \frameworkname outperforms existing popular methods including majority voting and self-debugging~\citep{chen2023teaching} (\cref{sec:exp_compare_methods}).
%     % \item \frameworkname not only consistently improves model capability on LiveCodeBench but also enhances performance on other benchmarks. (\cref{sec:exp_other_benchmark}).
%     \item \frameworkname consistently improves model capability on LiveCodeBench and other benchmarks (\cref{sec:exp_other_benchmark}).
% \end{enumerate}

% In addition, we include ablation studies on our selection method, the choice of hyper-parameters (i.e., number of parallel samples and temperature), the impact of incorporating in-context examples, and the effect of different revision methods in \cref{sec:ablations}.
% we introduce our experimental setup, the major benchmark results obtained using the default configuration in~\sect{sec:method}, and various ablation studies.

% and an ablation study on the effects of different selection methods, prompt optimization (\ssect{sec:scale_selection}).

\subsection{Experimental Setup}

\myparagraph{Models.} We consider both instruction-based and reasoning-based models. 
% In particular, we are interested in performance comparisons across models of different sizes with \frameworkname; therefore, we select a series of models within the same family. 
To compare performance across models of different sizes using \frameworkname, we select a series of models within the same family. 
We experiment with 12 models: (1) Instruction-based models: Qwen2.5-Coder-Instruct series (0.5B, 1.5B, 3B, 7B, 14B, 32B), \fouromini (0718 version)~\citep{hui2024qwen2, achiam2023gpt}; (2) Reasoning-based models: QwQ-32B-Preview, DeepSeek-R1-Distill-Qwen series (7B, 14B, 32B), and o1-mini~\citep{qwen_qwq_2024, guo2025deepseek, openai_learning_to_reason_2024}.

\myparagraph{Benchmarks.} 
We primarily use LiveCodeBench (MIT License) as our main evaluation benchmark, given its extensive usage by recent reasoning models and its inclusion of difficulty levels, which help analyze the behavior of different techniques~\citep{jain2024livecodebench,deepseek_r1_lite_2024,qwen_qwq_2024}. 
We use its v4 version for development (e.g., selecting hyper-parameters), which contains problems from August 2024 to November 2024.
% (e.g., in~\cref{fig:exp_hyper_parameters}). 
For final evaluation, we use v2 version that is non-overlapping to v4, and contain more problems. LiveCodeBench (v2) contains 511 problems, ranging from easy (182 problems), medium (206 problems), to hard (123 problems). 
In addition, we evaluate \frameworkname on CodeContests~\citep{li2022competition}, a collection of 165 challenging coding problems.
We use Pass@1 as our primary metric~\citep{chen2021evaluating}. Some experiments report Pass@N with N samples (often referred to as the `oracle' settings)~\citep{stroebl2024inference, brown2024large}.

\myparagraph{Baselines.} Our evaluation considers two categories of baselines. First, we assess our method's improvement over the original model (without test-time scaling), using three leading OpenAI reasoning models—o1-preview, o1-high, and o1-mini~\citep{openai_learning_to_reason_2024}—as performance benchmarks. Second, we evaluate different test-time scaling methods applied to the same models, encompassing both parallel (i.e., majority voting) and sequential (i.e., self-debugging) approaches.
% encompassing both parallel approaches (e.g., majority voting) and sequential approaches (e.g., self-debugging). \jx{Is this e.g., or i.e.,? we should be more specific about what baselines we are using.}
% Additionally, we provide more detailed comparisons with other selection strategies in ablation studies to show the effectiveness of our method.

\myparagraph{Implementation Details.}  We configure \frameworkname to generate 16 samples in parallel with a temperature of 0.7 (without top-p sampling) and perform 2 rounds of iterative debugging on public tests. We justify our choice of hyper-parameters in~\cref{sec:ablations}. Prompts are automatically generated by a prompting framework, DSPy, where detailed prompts can be found in~\cref{sec:appendix_prompts}~\citep{khattab2023dspy}. We run codes in a sandbox to avoid maliciously generated code, according to~\citep{chen2021evaluating}. Experiments with the largest model (DeepSeek-R1-Distill-Qwen32B) takes one day on 8 H100 GPUs. All experiments are conducted once. % Baselines are implemented with the same hyper-parameter: majority voting is implemented with 16 samples, and self-debugging is implemented by running 2 rounds of debugging from public tests.
% , with 16 samples used for the majority voting. 

%\textbf{Metric.} A recent work has pointed out that the overall performance is limited by the verifier’s ability~\citep{stroebl2024inference}. To disentangle such factors, we report the results of this section using an \textit{oracle verifier}, and without revision techniques.

% \subsection{~\frameworkname~effect on different models}
% \autoref{fig:all_models_performance} shows the performance comparison in LiveCodeBench with and without~\frameworkname, along with the o1-series reasoning models for comparison. We find that~\frameworkname~consistently improves model performance. For models in the same family, ~\frameworkname~consistently enables smaller models to outperform larger models: Qwen2.5-7B-Coder-Instruct, integrated with~\frameworkname~, outperforms Qwen2.5-32B-Coder-Instruct by 10.1\%. ~\frameworkname~also enables non-reasoning models to outperform reasoning models: \fouromini (0718), integrated with~\frameworkname, outperforms o1-preview. In addition, ~\frameworkname~consistetly improve strong reasoning models: the strongest open-source reasoning model, DeepSeek-R1-Distill-Qwen-32B, integrated with~\frameworkname, outperforms, o1-mini, and achieve near state-of-the-art result as o1 (high reasoning efforts).
\subsection{\frameworkname Main Results}
\label{sec:exp_main}
%\kurt{First, you need to make it clear throughout that in a single family, small models WITH our framework outperform large models WITHOUT our framework. 
%However, I think I would emphasize more the uniform improvement across all models and all families.
%Or, alternatively, describe an application where boosted performance on a smaller model enables use cases that would not otherwise be possible.}
\cref{fig:all_models_performance} presents a performance comparison on LiveCodeBench with and without \frameworkname, alongside the o1-series reasoning models for reference. Our results demonstrate that \frameworkname~consistently enhances model performance. 
When applied to models within the same family, \frameworkname allows small models to surpass large ones. For example, Qwen2.5-7B-Coder-Instruct integrated with \frameworkname outperforms Qwen2.5-32B-Coder-Instruct without \frameworkname by 10.1\%. 
Additionally, \frameworkname enables instruction-based models to surpass reasoning models, as evidenced by \fouromini (0718) with \frameworkname outperforming o1-Preview. Moreover, \frameworkname further improves strong reasoning models: the most capable open-source reasoning model, DeepSeek-R1-Distill-Qwen-32B, when enhanced with \frameworkname, surpasses o1-mini and achieves near state-of-the-art results comparable to o1 (high reasoning efforts). These results highlight that \frameworkname serves as a powerful test-time scaling technique that can effectively improve model performance across different scales, architectures, and reasoning capabilities.

\subsection{Comparison to Other Test-Time Methods}
\label{sec:exp_compare_methods}
% \xiuyu{Should we move this section to ablation studies, while moving \ssect{sec:ablation_selection} here instead as other selection methods are more like baselines than ablation of our method? On the other hand, investigating impacts of parallel, sequential, and integrated scaling is more like ablations.}
% We consider two paradigms for different test-time strategies: Parallel scaling based and sequential scaling based. For parallel scaling based, we consider majority voting based on the generated test inputs in this section, similar to the cluster method in~\citep{li2022competition}. For sequential scaling based paradigm, we consider self-debugging~\citep{chen2023teaching}. We use the same hyper-parameter for fair comparison: 16 parallel samples for majority voting, and 3 rounds for self-debugging. ~\autoref{tab:exp_diff_strategies} summarizes the performance. ~\frameworkname~consistently outperform either only using majority voting or using self-debugging.

% We evaluate \frameworkname against two different test-time scaling baselines of different paradigms:  \emph{parallel} and \emph{sequential} scaling approaches. 
We evaluate \frameworkname against two popular test-time scaling methods: majority voting~\citep{li2022competition} and self-debugging~\citep{chen2023teaching}. Majority voting employs parallel scaling: the model generates N samples, clusters them based on execution results~\citep{li2022competition}, selects the largest cluster, and randomly picks a final sample from it. Self-debugging follows a sequential scaling approach: the model generates a single sample, iteratively refines it using public tests~\citep{chen2023teaching}, and selects the final revised version.

To ensure fair comparison, we use consistent hyperparameters: 16 parallel samples for majority voting and 2 debugging rounds for self-debugging. \fouromini generates inputs for majority voting clustering and refines code samples for reasoning models. We use the model itself to refines code for non-reasoning models. As shown in~\cref{tab:exp_diff_strategies}, \frameworkname consistently outperforms both methods. For instance, for Qwen-2.5-Coder series, \frameworkname improves 8.4\% to 18.2\% to baselines. For the best performing model, DeepSeek-R1-Distill-Qwen-32B, \frameworkname outperforms the majority vote baseline by 9.9\%, and the self debugging baseline by 15.6\%. These results demonstrating the effectiveness of our hybrid approach.
% \jx{Can explain this result with some numbers if we want more text.}
% \xiuyu{need results to analyze whether parallel or sequential scaling helps more, if we want to discuss this.}
% \jx{The title of the upper figure. Non-reasoning to Instruction-based?}
% \xiuyu{Instruction-based sounds a bit distracting. Any issues with the name Non-reasoning?}

%\begin{figure}[h]
%    \centering
    % Placeholder figure
%    \includegraphics[width=0.99\linewidth]{latex/figures/compare_baselines.pdf}
%    \caption{\textbf{Pass@1 performance comparison between zero-shot, majority voting, self-debugging, and \frameworkname on LiveCodeBench(v2)}. Bold text denotes the best performance of the same model. "Qwen-Coder" is short for "Qwen2.5-Coder-Instruct", "R1-Distill" is short for "DeepSeek-R1-Distill-Qwen" . \frameworkname consistently outperforms other baselines. 
%    }
%    \label{fig:exp_diff_strategies}
%\end{figure}

\begin{table}[!t]
\centering
\resizebox{0.48\textwidth}{!}{%
\begin{tabular}{l|cccc}
\toprule
\textbf{Model} & Zero-Shot & S* & S* (Oracle) \\ 
\hline
Qwen-Coder-7B     & 1.8 & \textbf{10.9} (+9.1)  &  12.1 \\
Qwen-Coder-14B    & 9.7 & \textbf{21.8}  (+12.1) &  27.9 \\
Qwen-Coder-32B    & 10.1 & \textbf{21.8} (+11.7) & 29.7 \\
gpt-4o-mini       & 9.1 & \textbf{23.0} (+13.9)  & 28.5\\
\bottomrule
o1-mini       & 32.7 & \textbf{48.5} (+15.8) & 58.2\\
\bottomrule
\end{tabular}%
}
\caption{\textbf{Performance comparison on CodeContests}. 
Bold text denotes the best performance of the same model. 
"Qwen-Coder" is short for "Qwen2.5-Coder-Instruct", 
"R1-Distill" is short for "DeepSeek-R1-Distill-Qwen". 
\textit{\frameworkname consistently improves model performance on benchmark beyond LiveCodeBench.}}
\label{tab:codecontest}
\vspace{-5mm}
\end{table}

\subsection{Results on Other Benchmark}
\label{sec:exp_other_benchmark}

We further validate \frameworkname on CodeContests~\citep{li2022competition}. \cref{tab:codecontest} summarizes results, where \frameworkname consistently improves both instruction-based and reasoning models significantly. In particular, Qwen-2.5-Coder-7B-Instruct with \frameworkname improves 9.1\% from its zero-shot peformance of 1.8\%. It further outperforms \fouromini without \frameworkname by 1.8\%.

\section{Ablation Studies}
\label{sec:ablations}
% In this section, we conduct ablation studies to analyze the key components of \frameworkname. We examine the impact of different variations of iterative debugging (\cref{ssect:ablation_selfdebug}) and selection policies (\cref{sec:ablation_selection}). Additionally, we investigate how different hyper-parameter choices, such as the temperature setting and number of samples, affect parallel sampling performance (\cref{ssect:ablation_parallel}). We also assess the impact of further incorporating in-context example retrieval into the parallel sampling (\cref{sec:ablation_ICL}). All experiments in this study are conducted on LiveCodeBench (v4).
In this section, we conduct ablation studies to analyze the key components of \frameworkname, focusing on the effectiveness and variations within each stage of the framework. We evaluate the following aspects:
\begin{enumerate}
\item \textbf{Parallel Scaling:} We analyze the impact of different hyper-parameter choices, such as the temperature setting and the number of samples, on parallel sampling performance (\cref{ssect:ablation_parallel}). Additionally, we investigate the effect of incorporating in-context example retrieval into the parallel sampling process (\cref{sec:ablation_ICL}). We find that moderate temperatures improve performance, and adding ICL example can potential further improve performance.
\item \textbf{Sequential Scaling:} We explore variations of the iterative debugging process, including self-debugging with model-generated test cases (\cref{ssect:ablation_selfdebug}). We find that iteratively debugging from test execution feedback improve performance, even for reasoning models. We find that simply appending execution results from public tests for every iteration works the best.
\item \textbf{Selection Policy:} We assess the performance of different selection policies, comparing our adaptive input synthesis approach with alternative selection strategies (\cref{sec:ablation_selection}). We find that our adaptive input synthesis selection method is consistently more reliable than the generated tests and the LLM judge selection method.
\end{enumerate}
All ablation experiments are conducted on LiveCodeBench (v4).

\subsection{Parallel Sampling Hyper-Parameters}
\begin{figure}[h]
    \centering
    % Placeholder figure
    \includegraphics[width=.99\linewidth]{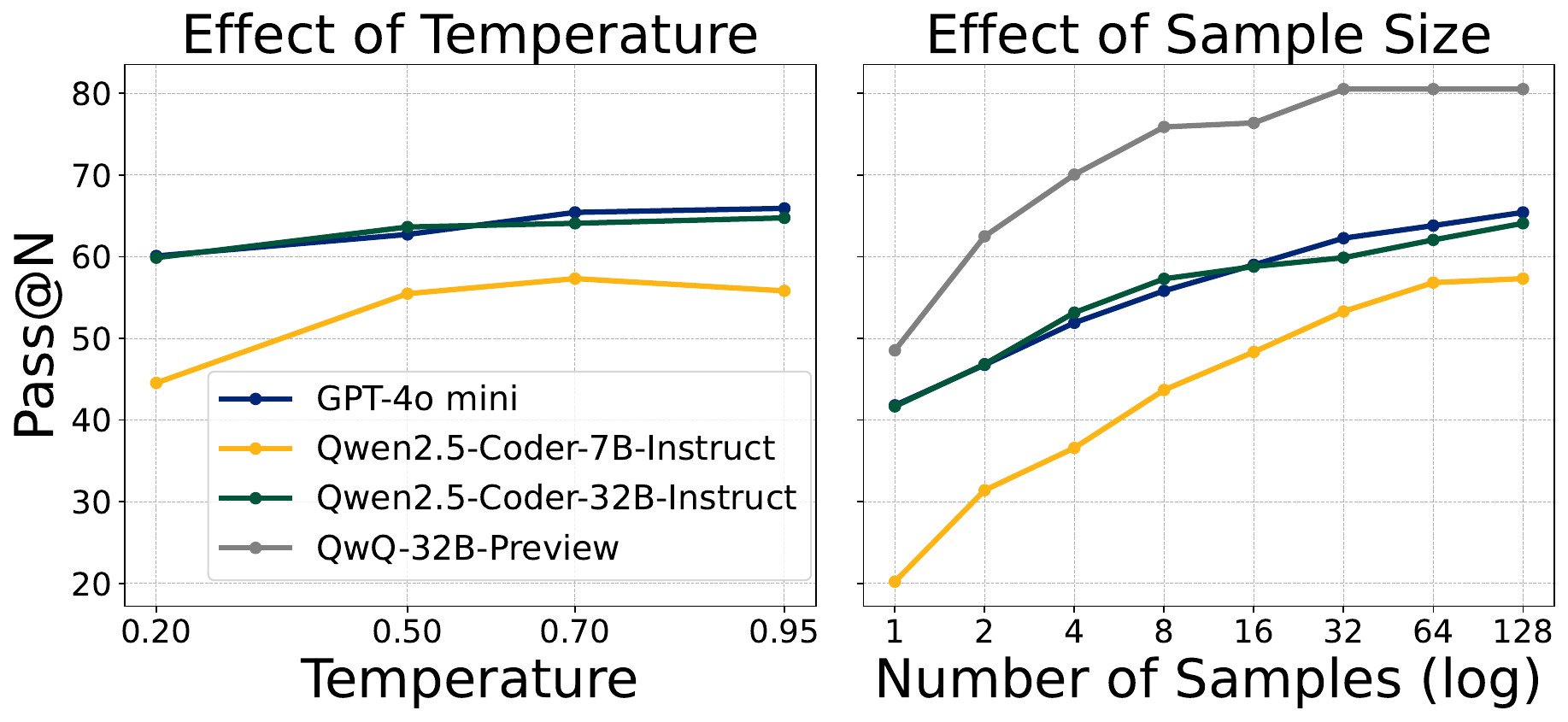}
    \caption{\textbf{The effect of hyper-parameters}. Left: The impact of temperature. A moderate temperature (0.7) balances diversity and quality, leading to higher Pass@N. In contrast, a higher temperature (0.95) does not further improve Pass@N, potentially degrading code quality. Right: The effect of increasing the number of samples. Performance improves log-linearly.}
    \label{fig:exp_hyper_parameters}
    % \vspace{-4mm}
\end{figure}
\label{ssect:ablation_parallel}

We examine the impact of two key factors in parallel sampling: temperature and the number of parallel samples. Understanding their influence is essential for optimizing test-time scaling strategies.

\myparagraph{Moderate temperatures improve performance, but high temperatures degrade it.}
%To analyze temperature effects, we vary  and report results in \cref{fig:exp_hyper_parameters} (left). 
\cref{fig:exp_hyper_parameters} (left) shows that moderate temperatures (0.2–0.7) enhance performance by balancing exploration and sample diversity. However, beyond 0.7, performance plateaus or declines, likely due to excessive randomness introducing noise. Some models, such as \qwensevenb, exhibit performance regression at higher temperatures, emphasizing the trade-off between diversity and solution consistency. These findings suggest that while moderate temperatures improve generation quality, excessively high values reduce code quality.

% For example, while \textit{\fouromini} attains a notable increase in \textbf{hard} task accuracy (from 2.04\% at 0.2 to 8.16\% at 0.95), the gains observed in medium-difficulty tasks do not follow an equally consistent upward trend. Indeed, some models experience performance regressions at higher temperatures, underscoring the trade-off between sampling diversity and solution consistency.

% Overall, these results reinforce the value of parallel sampling in improving coverage and final accuracy, highlighting (i) the substantial gains on harder tasks achieved by stronger models, (ii) the beneficial but nuanced effect of temperature.

\myparagraph{Repeated sampling improves performance, even for reasoning models.} 
% \myparagraph{Repeated sampling improves performance, even for reasoning models.} \autoref{fig:temp-various} (right) shows that, for \textbf{easy} problems, all models eventually reach nearly 100\% accuracy given sufficient samples (often by \(n=64\)). For instance, \textit{\fouromini} jumps from 80.7\% at \(n=1\) to 100\% by \(n=8\), and QwQ-32B-Preview achieves 100\% by \(n=2\). Meanwhile, on \textbf{medium} problems, stronger models exhibit larger gains, e.g., \textit{\fouromini} improves from 32.4\% at \(n=1\) to 70.3\% at \(n=128\), whereas a weaker model like \textit{Qwen2.5-Coder-7B-Instruct} moves from 8.1\% to 51.4\%. \textbf{Hard} problems benefit as well, albeit starting from low baselines; for example, \textit{Qwen2.5-Coder-32B-Instruct} rises from 2.04\% to 14.3\% by sampling 128 times. Interestingly, our preliminary experiments on a more recent reasoning-oriented model (\textit{QwQ-32B-Preview}) reveal a similar pattern: it scales from 4.08\% to 32.65\% on hard problems once sampled 64 times.
As shown in \cref{fig:exp_hyper_parameters} (right), increasing the number of parallel samples significantly improves performance across all models. Notably, \qwensevenb, the weakest performer at \( N=1 \), shows the largest gain, exceeding 35\% at \( N=64 \). Similarly, the more capable reasoning-model (QwQ-32B-Preview) follows the same trend, though its gains plateau beyond \( N=32 \). Nevertheless, it improves substantially, rising from 50\% at \( N=1 \) to 80\% at \( N=32 \). These results confirm that increasing the number of parallel samples is a simple yet effective strategy for enhancing performance in both instruction-following and reasoning-based models.

\subsection{Impact of In-Context Examples} 
\label{sec:ablation_ICL}

\begin{figure}[h]
    \centering
    % Placeholder figure
    \includegraphics[width=.95\linewidth]{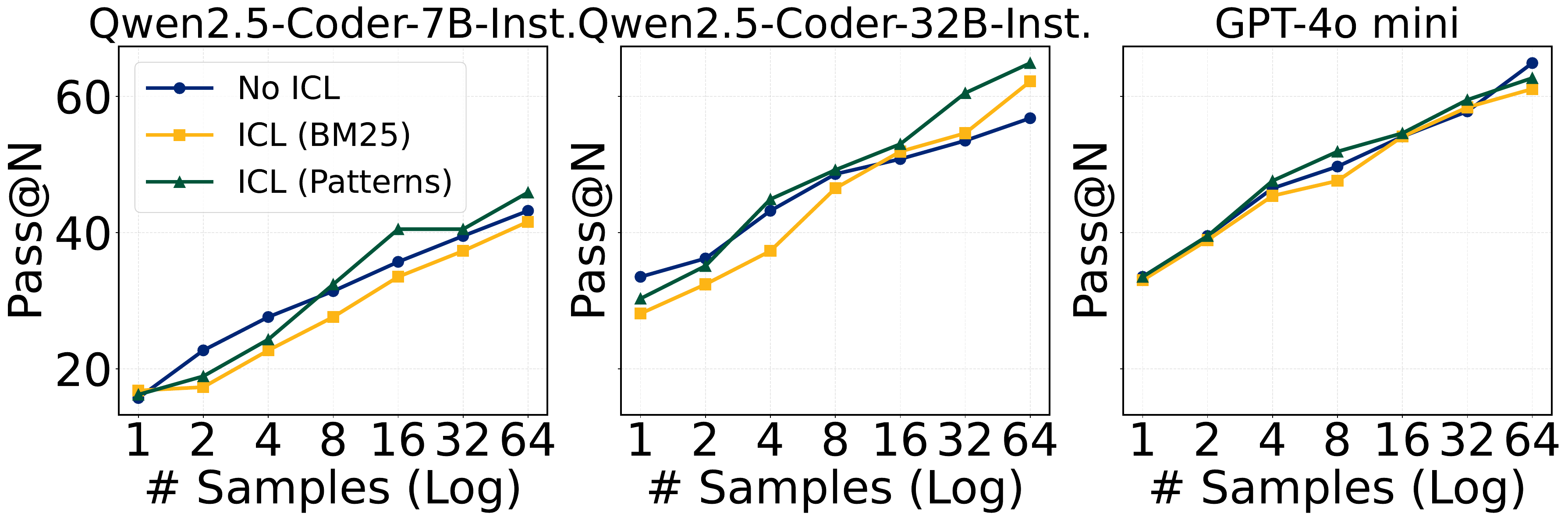}
    \caption{\textbf{Performance with in-context examples across different numbers of parallel samples (\(N\))}, for \fouromini, Qwen2.5-Coder-7B-Instruct, and Qwen2.5-Coder-32B-Instruct.}
    % \vspace{-5mm}
    \label{fig:icl_performance}
\end{figure}
% Using relevant in-context examples enhances generation diversity and performance.

While \frameworkname primarily focuses on repeated sampling for parallel scaling, it can be integrated with more advanced parallel scaling techniques. For instance, varying input prompts can create more diverse responses~\citep{lambert2024t}, which in turn may lead to better coverage.
In this ablation study, we investigate whether augmenting prompts with in-context examples can further improve parallel scaling performance.

% We investigate whether augmenting prompts with in-context examples enhances parallel scaling performance%\shiyi{add citation}
% , as existing works show varying input prompts can create more diverse responses~\citep{lambert2024t}.

We construct an example set from LiveCodeBench (v2) containing correct solutions and reasoning traces generated by GPT-4o mini. We explore two retrieval approaches for selecting in-context examples. \emph{ICL (BM25)} retrieves the top-$k$ similar prompts using a BM25 retriever and prepends each to a different sample when $n = k$~\citep{robertson2009probabilistic}. This approach is simple but may overlook solution-level similarities. \emph{ICL (Pattern)} groups problems by techniques (e.g., dynamic programming) and retrieves examples from the same technique, aiming to provide more relevant and structurally similar examples.

% The first, \emph{ICL (BM25)}, uses a BM25 retriever to fetch the top-$k$ similar prompts when $n = k$, and prepend each distinct one to a different sample. While straightforward, this method may overlook semantics and fail to capture solution-level similarities. To address this, our second approach, \emph{ICL (Pattern)}, categorizes problems based on recurring solution techniques (e.g., dynamic programming) and retrieves examples of the same type. This heuristic-driven, pattern-based method aims to provide more relevant and structurally similar examples. 

We evaluate medium-difficulty problems from LiveCodeBench (v4) with oracle selection. As shown in \cref{fig:icl_performance}, performance is highly sensitive to in-context example quality. ICL (BM25) performs similarly to or worse than the zero-shot baseline in most cases, except for $n=64$ with \qwenthirtytwob. In contrast, ICL (Pattern) outperforms the baseline when $n \geq 8$ for \qwensevenb and $n \geq 4$ for \qwenthirtytwob, while showing comparable performance with \fouromini.

These results highlight that selecting high-quality examples is crucial, and naive retrieval methods often fall short. Although ICL itself is promising, its performance is sensitive to example quality and retrieval effectiveness. We regard it as future work to develop robust ICL techniques that can be integrated into \frameworkname to further enhance parallel scaling performance.

% We conduct evaluation on medium-difficulty problems in LiveCodeBench (v4) with oracle selection. As shown in \cref{fig:icl_performance}, the choice of in-context examples significantly impacts performance. ICL (BM25) performs similarly to or worse than the zero-shot baseline in most settings, except for $n=64$ with \qwenthirtytwob. In contrast, ICL (Pattern) consistently outperforms the baseline when $n \geq 8$ for \qwensevenb and $n \geq 4$ for \qwenthirtytwob, while performing comparably with \fouromini across all $n$ samples. 
% This underscores that retrieving the right in-context examples is both challenging and crucial—simple methods often fall short. Notably, the right examples could improve parallel sampling performance, especially when $n$ is large. Although we choose not to incorporate them into our final iterative debugging solution to avoid additional complexity, the results point toward future directions. In particular, building a larger, more diverse, and more relevant ICL example database and developing better retrieval approaches show promise to further enhance performance.

% Overall, these results reinforce the value of parallel sampling in improving coverage and final accuracy, highlighting (i) the substantial gains on harder tasks achieved by stronger models, (ii) the beneficial but nuanced effect of temperature, and (iii) the complexity and potential limits of simple in-context approaches in further boosting parallel sampling.

\subsection{Impact of Iterative Debugging Variants}
\label{ssect:ablation_selfdebug}
\begin{figure}[h]
    \centering
    % Placeholder figure
    \includegraphics[width=.99\linewidth]{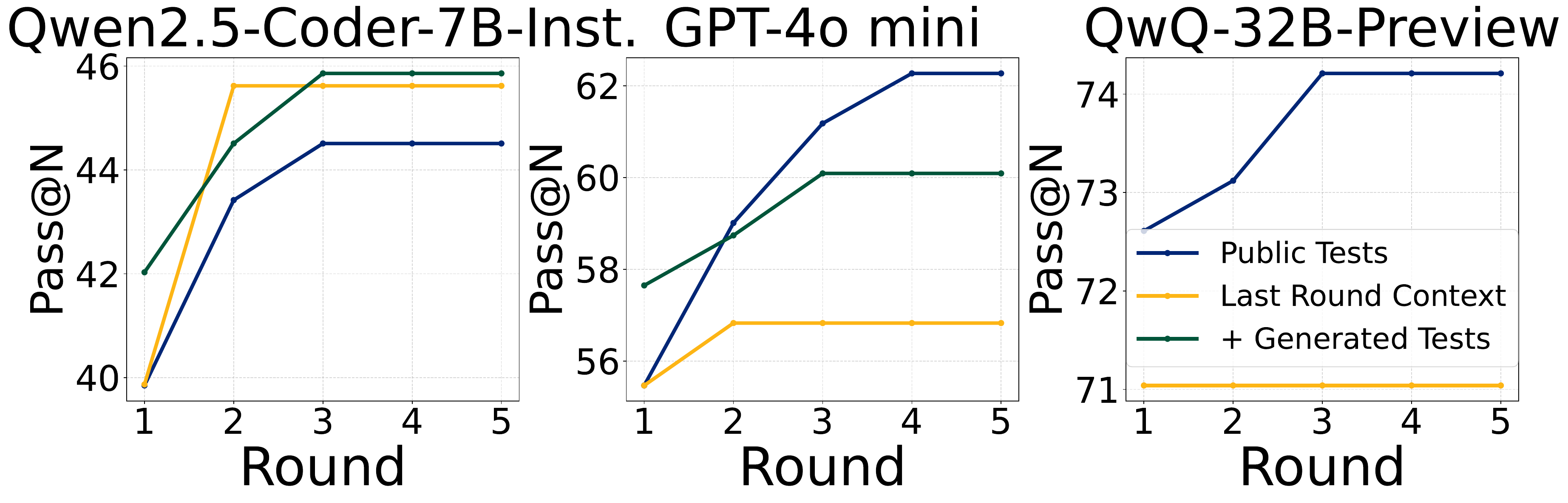}
    \caption{\textbf{Comparison of three iterative debugging approaches}: \textit{Public Tests},  \textit{+ Generated Tests}  and \textit{Last Round Context}. Results are obtained with \(N=8\), \(\text{temperature}=0.7\) and up to four rounds of debugging.}
    \label{fig:iterative_debug_performance}
    \vspace{-5mm}
\end{figure}
We examine the effectiveness of three variants of iterative debugging: (1) \textbf{Public Tests}: The model iteratively debugs using public tests and stops once the sample passes all of them. (2) \textbf{+Generated Tests}: In addition to public tests, the model continues debugging on model-generated tests following the algorithm in~\citep{ridnik2024code}. (3) \textbf{Last Round Context}: The model iteratively debugs using only public tests, but instead of using code samples from all previous rounds for debugging, it uses only the last round of code sample as context. This is motivated by observations that LLMs may perform sub-optimally when handling large context windows~\citep{liu2024lost}.

\cref{fig:iterative_debug_performance} summarizes the result. We find that: (1) \textit{Even though reasoning models implicitly perform self-reflection and revising, they benefit from explicit debugging through test execution feedback}: the performance of QwQ-32B-Preview model improves from 72.6 to 74.2 with 2 rounds of debugging. 
(2) \textit{Reducing the context window or considering more model-generated tests does not show consistent improvement}: while using only the last round of context improves performance for the Qwen2.5-Coder-7B-Instruct model, it results in worse performance for the other two models. Similarly, incorporating additional model-generated tests does not enhance performance for \fouromini.
(3) \textit{The benefits of iterative debugging tend to plateau, typically after 2–3 rounds}: this finding aligns with the observation that the benefit of sequential scaling flattens out~\citep{muennighoff2025s1}. Motivated by these findings, we choose to use 2 round of debugging, only on public tests for simplicity, and apply iterative debugging even for reasoning models in~\cref{sec:exp_main}.

\subsection{Impact of Different Selection Policies}
\label{sec:ablation_selection}
\begin{table}[h]
\centering
\resizebox{0.48\textwidth}{!}{%
% \begin{tabular}{l|cc>{\columncolor{lightblue}}c}
\begin{tabular}{l|cccc}
\toprule
\textbf{Model} & Public & Generated & LLM & Adaptive Input\\ 
& Only & Tests & Judge & Synthesis (Ours)\\ 
\hline
Qwen-Coder-7B     &  42.3 & 42.3  & 42.3 &  \textbf{44.5} \\
Qwen-Coder-32B    &  54.6 & 57.8  & 55.5 &  \textbf{58.3} \\
\fouromini       &  54.1 &  55.2 & 56.3 & \textbf{57.3} \\
QwQ-32B-Preview & 64.3 & 65.9 & 68.6 &  \textbf{69.7}\\
\rowcolor{black!10} Avg. & 53.8 & 53.1 & 55.6 & \textbf{57.5}\\
\bottomrule
\end{tabular}%
}
\caption{\textbf{Pass@1 Performance comparison between different selection methods on LiveCodeBench(v4).} 
Bold text denotes the best performance of the same model. 
"Qwen-Coder", "R1-Distill" is short for "Qwen2.5-Coder-Instruct" and "DeepSeek-R1-Distill-Qwen". Results are obtained with N=8 and temperature=0.7. \textit{Our Adaptive Input Synthesis method achieves better accuracy over other methods.}}
\label{tab:diff_selection}
\vspace{-2mm}
\end{table}
We compare different policies for selecting the best sample after iterative debugging. We evaluate four approaches: 
% (1) \textbf{Random}: randomly selecting one sample after debugging; \xiuyu{"after debugging" sounds misleading since random doesn't benefit from debugging? Also table 2 does not contain random -- should be removed here if we don't want to include it.} 
(1) \textbf{Public Only}: using only public test cases for selection and randomly selecting a sample if it passes all tests; (2) \textbf{Generated Tests}: applying public test filtering followed by additional test case generation using \fouromini, selecting the sample that passes the most test cases; (3) \textbf{LLM Judge}: applying public test filtering and then using LLMs for pairwise selection among code samples; and (4) \textbf{Adaptive Input Synthesis} —applying the selection algorithm described in~\ssect{sec:scale_selection} with \fouromini after public test filtering. % Our experiments are conducted using 8 parallel samples with a temperature of 0.7 in LiveCodeBench v4.

% ~\autoref{tab:exp_diff_strategies} summarizes the performance. We finds that firstly, selection is important, even for strong model after iterating on several rounds of self-debugging: For instance, o1-mini benefits from simple public test filtering by 8\%, even though these tests are available during its self-debugging procedure. Secondly, neither LLM-as-a-judge or generated tests can relaibly generate better selection than the available public tests. In contrast, the test-assisted LLM-as-a-judge can provide stable improvement. 
\cref{tab:diff_selection} summarizes the results. Notably, the Generated Tests approach does not yield improvements over the Public Only baseline. 
This is due to errors in model-generated outputs, which, when applied to poorly chosen inputs, introduce significant noise in the selection process, ultimately degrading performance. 
In contrast, our Adaptive Selection method enables the LLM to strategically select an input that best differentiates samples while avoiding the need to predict outputs. By leveraging real execution outputs rather than model predicttions, the LLM makes more reliable decisions, leading to improved selection accuracy. 
%  Our findings indicate that selection plays a crucial role, even for strong models after multiple rounds of self-debugging. For example, o1-mini benefits from simple public test filtering with an 8\% performance improvement, despite these tests being available during its self-debugging process. Furthermore, neither LLM-as-a-Judge nor generated tests consistently outperform selection based on public tests. In contrast, our test-assisted LLM-as-a-Judge approach provides significantly more stable and reliable improvements.  

\section{Conclusion}
%\kurt{I would summarize quantitative improvements here as you did in the abstract.}
% We propose~\frameworkname, a simple and effective framework to scale test-time compute for code generation. It integrates iterative debugging methods into the existing repeated sampling based paradigm, together with \textit{adaptive input synthesis} that selects the final sample with grounded information. ~\frameworkname has been shown to consistently improve code generation capability on LiveCodeBench and CodeContests. With~\frameworkname, instruction-tuned model outperforms reasoning models: \fouromini with our method outperforms o1-preview without our method by 3.7\% on LiveCodeBench); open reasoning model can further match closed sourced state-of-the-art: DeepSeek-R1-Distill-Qwen-32B with our method achieves 86.7\% on LiveCodeBench, close to o1 (high reasoning efforts) without our method at 88.5\%.
We propose \frameworkname, the first hybrid test-time scaling framework for code generation that substantially improves both coverage and selection accuracy. \frameworkname extends the existing parallel scaling paradigm with sequential scaling through iterative debugging and incorporates \textit{adaptive input synthesis}, a novel mechanism that synthesizes distinguishing test inputs to differentiate candidates and identify correct solutions via execution results.

\frameworkname consistently improves code generation performance across benchmarks, including LiveCodeBench and CodeContests. Notably, \frameworkname enables a 3B model to outperform \fouromini, \fouromini to surpass o1-preview by 3.7\% on LiveCodeBench, and DeepSeek-R1-Distill-Qwen-32B to achieve 86.7\% on LiveCodeBench, approaching o1-high at 88.5\%.
% (1) instruction-tuned models to outperform reasoning models: \fouromini with \frameworkname surpasses o1-preview without \frameworkname by 3.7\% on LiveCodeBench; and (2) open reasoning models to match the performance of state-of-the-art closed-source models: DeepSeek-R1-Distill-Qwen-32B with \frameworkname achieves 86.7\% on LiveCodeBench, approaching o1-high without \frameworkname at 88.5\%.
\section{Limitations}
This work primarily focuses on competition-level code generation, where it does not studies tasks such as software engineering tasks, e.g., SWE-BENCH~\citep{jimenez2023swe}. The method primarily focuses on improving accuracy, while it does not aim for minimizing costs.
\section{Acknowledgment}
This work is funded by the Sky Computing Lab at UC Berkeley. We extend our sincere gratitude to Matei Zaharia and Anastasios Nikolas Angelopoulos for their invaluable feedback. We are grateful for the generous compute resources support from Databricks, Lambda Labs, and Anyscale. In particular, we thank Jonathan Frankle (Databricks) and Chuan Li (Lambda Labs) for facilitating access to these resources.
% Bibliography entries for the entire Anthology, followed by custom entries
%\bibliography{anthology,custom}
% Custom bibliography entries only
\bibliography{custom}

\appendix

% \section{Example Appendix}
\section{Appendix}
\label{sec:appendix}

\subsection{Example of Coding Problem}
\label{sec:appendix_example_code}
In the method section (\cref{sec:method}), we introduce our problem setup, which includes unambiguous configuration with a small amount of demonstrations. In this section, we provide one such example to better illustrate how typically dataset provide questions. In particular, we show one sample from the hard subset of LiveCodeBench~\citep{jain2024livecodebench}.

\begin{tcolorbox}
[colback=white,colframe=gray,title=Question]
You are given a string word and an array of strings forbidden.
A string is called valid if none of its substrings are present in forbidden.
Return the length of the longest valid substring of the string word.
A substring is a contiguous sequence of characters in a string, possibly empty.
\newline

Example 1:

Input: word = "cbaaaabc", forbidden = ["aaa","cb"]

Output: 4

Explanation: There are 11 valid substrings in word: "c", "b", "a", "ba", "aa", "bc", "baa", "aab", "ab", "abc" and "aabc". The length of the longest valid substring is 4. 
It can be shown that all other substrings contain either "aaa" or "cb" as a substring. 
\newline

Example 2:

Input: word = "leetcode", forbidden = ["de","le","e"]

Output: 4

Explanation: There are 11 valid substrings in word: "l", "t", "c", "o", "d", "tc", "co", "od", "tco", "cod", and "tcod". The length of the longest valid substring is 4.
It can be shown that all other substrings contain either "de", "le", or "e" as a substring. 
\newline
 
Constraints:

1 $\leq$ word. length $\leq 10^5$ word consists only of lowercase English letters. 1 $\leq$ forbidden. length $\leq 10^5$. 1 $\leq$ forbidden$[i]$. length $\leq$ 10. forbidden[i] consists only of lowercase English letters.

\label{box:example_code}
\end{tcolorbox}

\subsection{Prompt templates}
\label{sec:appendix_prompts}
\begin{figure*}[t]
\centering
\includegraphics[width=\textwidth]{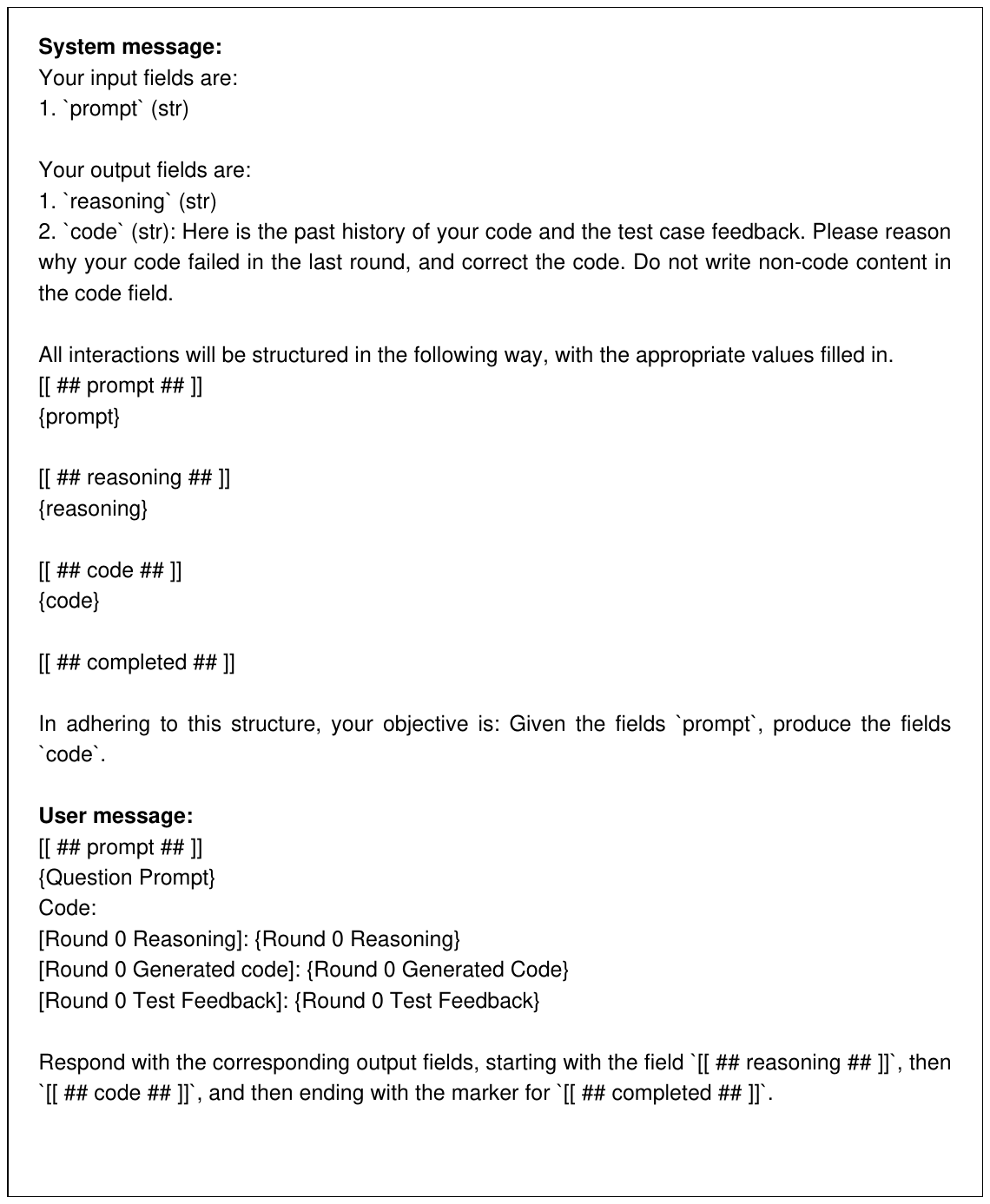}
\caption{The prompt for iterative debugging.}
\label{fig:self-refine}
\end{figure*}

\begin{figure*}[t]
\centering
\includegraphics[width=\textwidth]{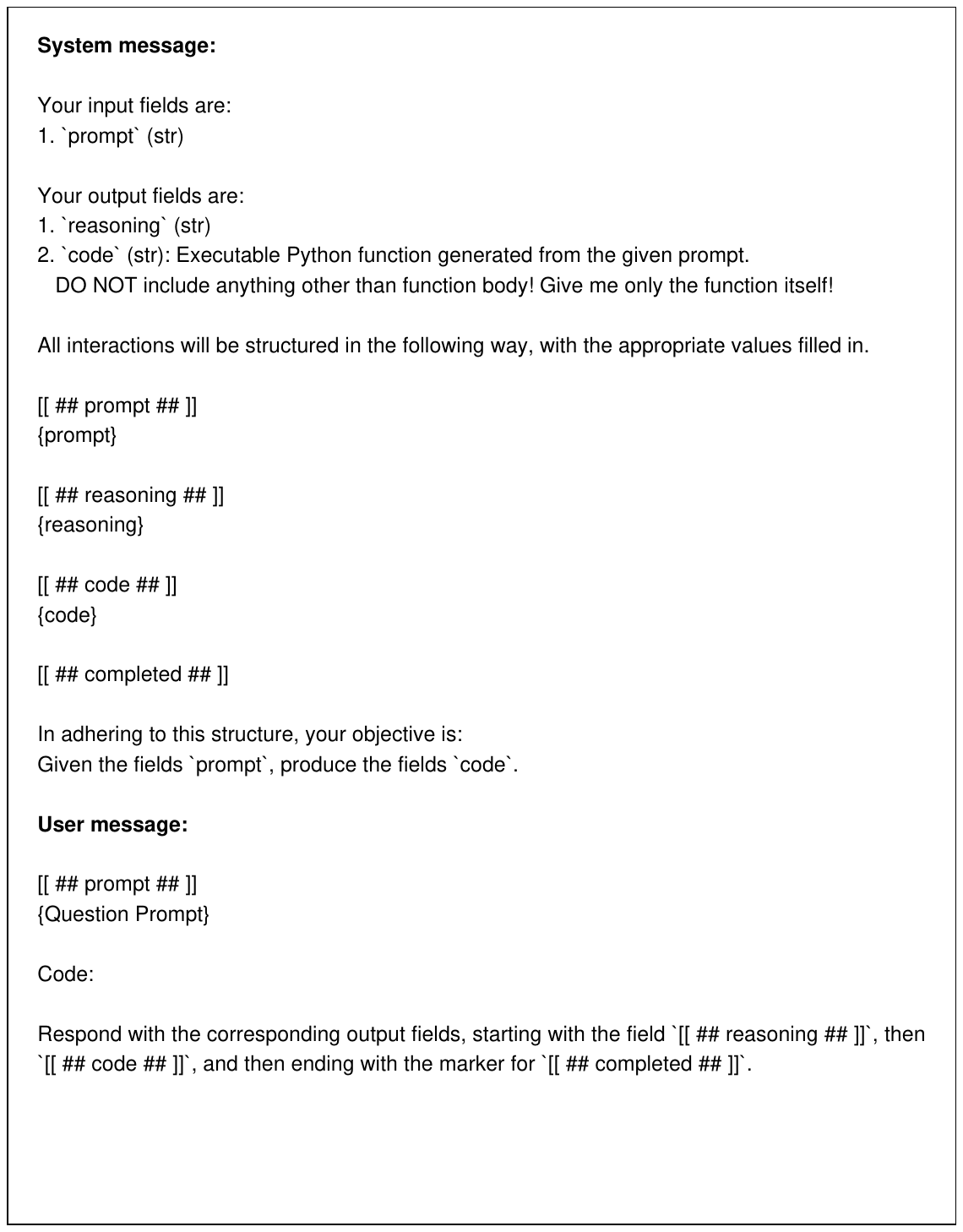}
\caption{The prompt for code generation.}
\label{fig:generation-prompt}
\end{figure*}

We also provide detailed prompts used in our experiments in \cref{fig:self-refine} to \cref{fig:generation-prompt}. These prompts are generated automatically by DSPy~\citep{khattab2023dspy}.
% Please refer to our github repository~\footnote{\url{} for full details.

\begin{figure*}[t]
\centering
\includegraphics[width=\textwidth]{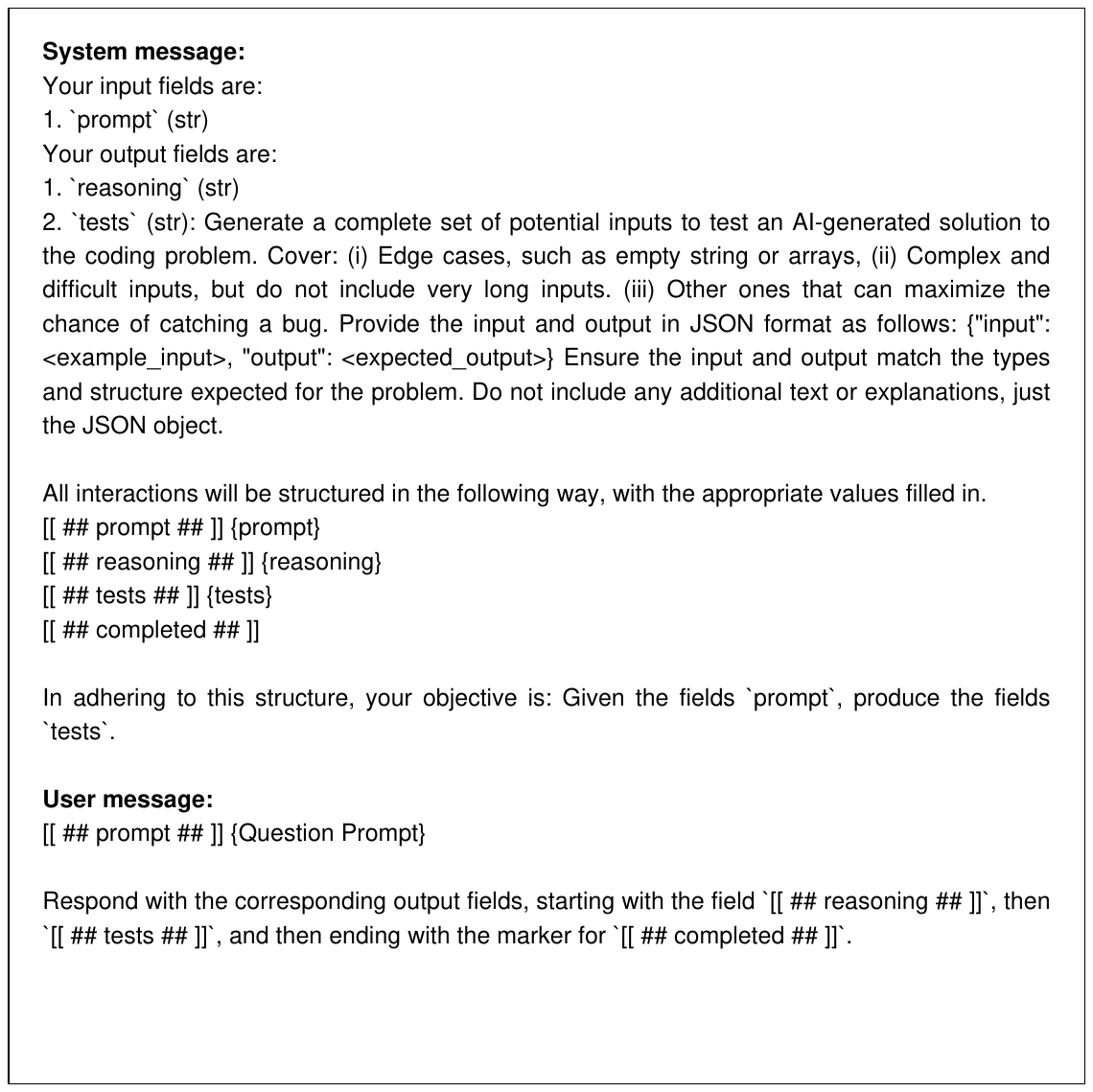}
\caption{The prompt for generating test cases.}
\end{figure*}

\end{document}